\documentclass[sigconf, language=french,
language=german, language=spanish, language=english]{acmart}

\usepackage{booktabs}
\usepackage{array}
\usepackage{tabularx} 
\usepackage{amsmath}
\AtBeginDocument{%
  \providecommand\BibTeX{{%
    \normalfont B\kern-0.5em{\scshape i\kern-0.25em b}\kern-0.8em\TeX}}}

\setcopyright{acmlicensed}
\acmDOI{10.1145/3664646.3665084}
\acmYear{2024}
\copyrightyear{2024}
\acmSubmissionID{fsews24aiwaremain-p27-p}
\acmISBN{979-8-4007-0685-1/24/07}
\acmConference[AIware '24]{Proceedings of the 1st ACM International Conference on AI-Powered Software}{July 15--16, 2024}{Porto de Galinhas, Brazil}
\acmBooktitle{Proceedings of the 1st ACM International Conference on AI-Powered Software (AIware '24), July 15--16, 2024, Porto de Galinhas, Brazil}
\received{2024-04-24}
\received[accepted]{2024-05-10}

\usepackage{amsmath}

\begin{document}

\title{Investigating the Potential of Using Large Language Models for Scheduling}

\author{Deddy Jobson}
\email{deddy@mercari.com}
\affiliation{%
  \institution{Mercari}
  \city{Tokyo}
  \country{Japan}
}

\author{Li Yilin}
\email{y-li@mercari.com}
\affiliation{%
  \institution{Mercari}
  \city{Tokyo}
  \country{Japan}
}







\renewcommand{\shortauthors}{Jobson and Li}

\begin{abstract}
The inaugural ACM International Conference on AI-powered Software introduced the AIware Challenge, prompting researchers to explore AI-driven tools for optimizing conference programs through constrained optimization. We investigate the use of Large Language Models (LLMs) for program scheduling, focusing on zero-shot learning and integer programming to measure paper similarity.
Our study reveals that LLMs, even under zero-shot settings, create reasonably good first drafts of conference schedules. When clustering papers, using only titles as LLM inputs produces results closer to human categorization than using titles and abstracts with TFIDF. The code has been made publicly available\footnote{https://github.com/deddyjobson/llms-for-scheduling}.
\end{abstract}




\received{20 February 2007}
\received[revised]{12 March 2009}
\received[accepted]{5 June 2009}

\maketitle

\section{Introduction} 
The inaugural ACM International Conference on AI-powered Software debuts the AIware Challenge, encouraging researchers and practitioners to implement AI-driven tools to create an optimized conference program, a fundamental issue in constraint optimization. Constrained optimization is a prevalent mathematical task, needing intensive computational resources and tailored algorithms to address the unique complexities of the decision space. Applying Large Language Models (LLMs) has demonstrated significant success in various domains such as mathematical problem-solving and reasoning.

This report investigates the feasibility of leveraging LLMs to address the program scheduling challenge. Our investigation revolves around two primary approaches: 1) zero-shot learning to generate the schedules and 2) integer programming with LLMs to measure the similarity between papers.


\section{Problem Statement} 

Our paper focuses on automating the allocation of papers to predetermined sessions, treating it as an allocation and constrained clustering problem. We do not factor in the specific timing of paper presentations or the impact of parallel tracks, earmarking these as potential areas for future investigation on the broader goal of automating conference scheduling.



\section{Experiments} 


\subsection{Zero-Shot Scheduling by LLMs} 
\label{subsec:zero-shot}
Our first AIware approach directly prompts a large language model to make the schedule. While LLMs have a lot of room to improve on arithmetic reasoning tasks, recent approaches have been proposed to improve their abilities for math problems\cite{yang_large_2023,zhang_-context_2024,levonian_retrieval-augmented_2023}. Furthermore, LLMs are expected to improve at solving LLMs at zero-shot configurations, too, due to the scaling laws\cite{wei_emergent_2022}. 

However, from our experiments, we find the LLMs are still unable to create schedules without violating constraints. We, therefore, additionally experiment with smaller-scale problems by downsampling the number of sessions or papers within each session. To measure the performance of our approach, we measure the following metrics:
\begin{itemize}
    \item Completeness and Homogeneity wrt the ground truth\cite{rosenberg2007v}
    \item \% of constraints broken, for various types of constraints. 
\end{itemize}
We use a temperature setting of 0.8 in our experiments. 

We find that GPT-4 in zero-shot settings can generate reasonably good conference schedules. We also counted the number of violations according to the schedule created by the GPT -4. Looking at violations on the number of sessions and papers in the proposed schedule, we find that, on average, less than 3 papers were missing, and two sessions were added. Furthermore, checking if the session durations are within bounds, we find that while many sessions proposed were too long (in some cases, more than 50\%), in most cases, the proposed session was off by less than 10\% of the length of the session. This suggests that by moving around a couple of papers between sessions, a human can finish the work mostly done by an LLM, suggesting a collaboration with LLMs be the best approach. 

\begin{figure}
    \centering
    \includegraphics[width=1\linewidth]{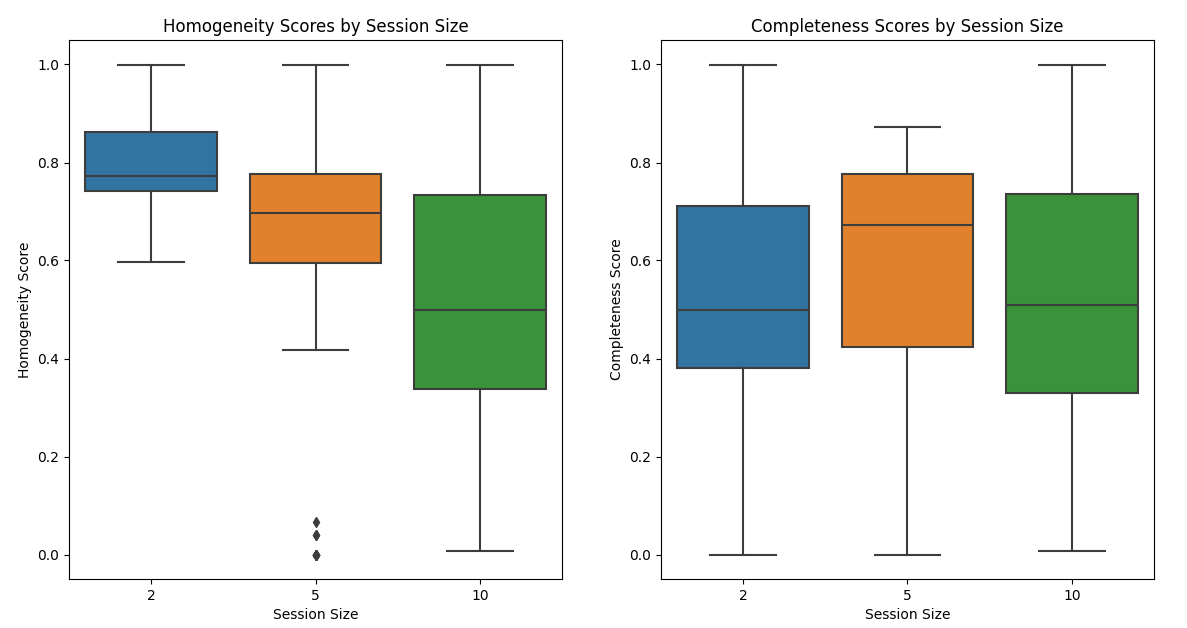}
    \caption{Comparing the homogeneity and completeness score for different number of papers per session. Homogeneity deteriorates when the scale of the data is larger, but not completeness.}
    \label{fig:corrected_completeness}
\end{figure}

Computing the homogeneity and completeness scores with respect to the actual schedule of MSR 2022 (Figure \ref{fig:corrected_completeness}), we find that while homogeneity drops with an increase in problem size, completeness is not as much affected.

\subsection{Integer Program using LLM to measure similarity}
In Section \ref{subsec:zero-shot}, we discover that LLMs face challenges in strict adherence to all constraints, particularly when dealing with many papers. We plan to explore how to leverage the text-understanding capabilities of LLMs to facilitate the problem. 

We incorporate the similarity of papers as an objective in our optimization process to build an integer programming problem and use LLM to generate the similarity. Recognizing that similarity is a subjective concept, we have utilized the existing session arrangement as the benchmark for comparison. To assess the proximity of our results to the original schedule, we employ the completeness score and homogeneity score metrics \cite{rosenberg2007v}.

Leveraging LLMs' text-understanding capabilities, we cluster the papers into 5 groups and contrast the outcomes with a Bag of Words approach featuring TFIDF normalization. The comparison, outlined in Table \ref{table:clustering_result_comparison}, demonstrates the superior performance of LLM when utilizing only the paper titles compared to TFIDF with both titles and abstracts.

To further evaluate the clustering outcomes, we formulate an integer programming problem (formula see Appendix) using the best results from LLM with text alone and TFIDF with text and abstract as the similarity parameters. The similarity metric indicates that papers $i$ and $j$ sharing the same cluster exhibit a similarity value of 1, and 0 otherwise. completeness score and homogeneity score are outlined in Table \ref{table:optimization_clustering_result_comparison}. Notably, the comprehensive and homogeneity scores align LLM's clustering result closely with TFIDF. This alignment reflects a common scenario in optimization problems where the parameters are derived from predicted outcomes.

\begin{table}[h] 
    \centering
    \begin{tabular}{|>{\raggedright\arraybackslash}p{3cm}| >{\raggedleft\arraybackslash}p{2cm} | >{\raggedleft\arraybackslash}p{1.8cm}|} 
        \toprule 
        \textbf{Approaches} & \textbf{Completeness\newline Score} & \textbf{Homogeneity\newline Score} \\ 
        \midrule 
        TFIDF (title only) & 0.33 & 0.14 \\ 
        TFIDF (title and abstract) & 0.37 & 0.20 \\
        LLM (title only) & 0.41 & 0.22 \\ 
        \bottomrule 
    \end{tabular}
    \caption{clustering results applying LLM and TFIDF. This is the average results from 5 random trials. LLM with title only exhibits the highest performance in completeness score and homogeneity score when considering the current session label as the ground truth.} 
    \label{table:clustering_result_comparison} 
\end{table}
\begin{table}[h] 
    \centering
    \begin{tabular}{|>{\raggedright\arraybackslash}p{3cm}| >{\raggedleft\arraybackslash}p{2cm} | >{\raggedleft\arraybackslash}p{1.8cm}|}
        \toprule
        \textbf{Approaches} & \textbf{Completeness\newline Score} & \textbf{Homogeneity\newline Score} \\
        \midrule
        TFIDF (title and abstract) & 0.42 & 0.42 \\
        LLM (title only) & 0.41 & 0.41 \\
        \bottomrule
    \end{tabular}
    \caption{The results of integer programming optimization using similarity measures from LLM and TFIDF. Due to the inferior performance of TFIDF with title-only, it is omitted from this comparison. The results show that the LLM has a similar result to the TFIDF(title and abstract).} 
    \label{table:optimization_clustering_result_comparison} 
\end{table}

\section{Conclusion}
Our experimentation shows that LLMs are a promising tool in managing optimization problems even when they lack precision when the number of decision variables exceeds a certain threshold. To address this limitation, a potential strategy involves involving humans in the loop, or combining the strengths of LLMs and numerical solvers as we demonstrated through our clustering + Integer Programming approach.

\section{Acknowledgements}
We would like to thank Mercari Inc. for supporting this research. LLMs were used to reword sections and to reformat tables in the paper. However, we did not generate entire sections through LLMs. 

\bibliographystyle{ACM-Reference-Format}
\bibliography{references-deddy,references_yilin} 

\appendix
\section{Prompt Design}
The following is the template of the prompt we used. Note that the terms surrounded by curly braces (\{\}) show where the list of papers was added. 

\begin{verbatim}
Program creation is the process of taking all the 
accepted papers to a conference and allocating a 
presentation slot for each paper with parallel sessions. 
The PC chairs of a conference typically do this manually. 
Assign sessions to the following papers based on
the following constraints:
1. The total length of all paper presentations 
   within a session should be less than or equal to 
   the length of the session they are in. 
2. No new sessions should be added. 
3. The "Discussions and Q/A" event should only be 
   at the end of a session if present. 
4. All papers must be assigned to some session. 

The output should contain the schedule in the form 
of the csv representation of a data frame. This 
csv representation should be in three quotes (```) 
on both sides so that I can easily extract it 
from your result and make a data frame. 
Example output format(use as many rows as the 
actual number of papers):

```
session@talk_title@duration
231@An Empirical Study on Maintainable Method ...@7
223@Improve Quality of Cloud Serverless ...@7
15@Extracting corrective actions from code ...@7
15@How to Improve Deep Learning for Software ...@7
11@ReCover: a Curated Dataset for Regression...@4
```

The list of paper and session titles are below: 

Session Lengths:
{sessions_df_string}

Paper durations:
{papers_df_string}


Make sure, above all else, that your response 
is formatted as requested, with the proper headers.
\end{verbatim}

Some points to note: 
\begin{itemize}
    \item On top of giving the list of papers, we gave a list of sessions with session length. 
    \item We shuffled the list of papers so the LLM would not copy the original ordering. 
    \item We opted for values separated by the commercial at symbol (@) rather than commas to make handling titles with commas easier. 
    \item The final statement was added to increase the chances of getting the output in the right format. 
\end{itemize}

\section{Integer Programming Formulation}
Define binary decision variables \( x_{i,j} \):
\[ x_{i,j} = 
\begin{cases} 
1 & \text{if paper } i \text{ is scheduled in session } j, \\
0 & \text{otherwise}.
\end{cases} \]
We introduce a new variable to indicate whether two papers are scheduled for the same session. Define binary decision variables \( z_{i,j,m} \):
\[ z_{i,j,m} = 
\begin{cases} 
1 & \text{if papers } i \text{ and } j \text{ are in the same session } m, \\
0 & \text{otherwise}.
\end{cases} \]

\subsection{Objective Function}
Maximize the similarity between papers scheduled in the same session (if the paper is clustered into the same cluster, the similarity is 1 otherwise it is 0):
\[ \text{Maximize} \quad \sum_{i=1}^{N}\sum_{j=1}^{N} \sum_{m=1}^{M} \text{simlarity}{(i,j)} \cdot z_{i,j,m} \]

\subsection{Constraints}
\subsubsection{Link \( z \) and \( x \) variables}
For each pair of papers both scheduled in the same session:
\[ z_{i,j,m} \geq x_{i,m} + x_{j,m} - 1 \]
\[ z_{i,j,m} \leq x_{i,m} \]
\[ z_{i,j,m} \leq x_{j,m} \]
\subsubsection{Each paper scheduled exactly once}
Ensure every paper is scheduled in exactly one session:
\[ \sum_{j=1}^{M} x_{i,j} = 1 \quad \forall i \]
\subsubsection{Session length constraints}
Ensure the total duration of papers in any session does not exceed the available time:
\[ \sum_{i=1}^{N} x_{i,j} \cdot \text{papers\_duration}_i \leq \text{session\_lengths}_j \quad \forall j \]

\end{document}